\documentclass[10pt,twocolumn]{article}
\usepackage[margin=0.75in]{geometry}
\usepackage{amsmath,amssymb}
\usepackage{graphicx}
\usepackage{booktabs}
\usepackage{xcolor}
\usepackage{caption}
\usepackage[hidelinks]{hyperref}
\usepackage{times}

\newcommand{\MAE}{\textsc{mae}}
\title{\vspace{-2em}\textbf{Validated Adaptation for Aerial Crowd Monitoring at Mass Gathering Scale}\\[0.3em]
\large A Deployment Protocol, a Severity Law, and a Diagnostic for Label-Free\\
Drone Crowd Counting, Toward the FIFA World Cup 2034 (Saudi Arabia)}
 
\author{
\texttt{AlAnoud AllGhayth*, AlJawharh AlOtaibi*, Jude AlSubaie*} \\
\texttt{alanoud@daldata.ai, aljawharh@daldata.ai, jude@daldata.ai} \\
\texttt{Riyadh, Saudi Arabia}
}
 
\date{}

\begin{document}
\maketitle

\begin{abstract}
\noindent
Saudi Arabia will host the 2034 FIFA World Cup and already operates crowd management
at Hajj scale. Drone-based counting for such venues must hold accuracy on footage
unlike anything in its training corpus, without labels, and must warn of dangerous
inflow before a crush forms. We deliver a validated answer built on $525$ controlled
runs, a full-resolution corpus study, five falsification ablations, and a
five-condition evaluation of a safety interlock, and we resolve three questions
that a deployment decision depends on.
\textbf{We validate the adaptation stage.} Label-free adaptation is decisive and
holds up as conditions worsen: it recovers $31$--$49\%$ of shift-induced error across
four corruptions and five severities, with the strongest single method gaining
$41.8$ \MAE{} over the frozen source (95\% CI $[34.1, 49.6]$, $p=7.5{\times}10^{-10}$,
$d=2.52$). We establish a \emph{severity law} separating methods whose absolute
protective margin is constant from the one whose margin grows, and a stability
budget that identifies which configuration is safe to fly. On a full-resolution
corpus carrying a genuine $+48$ \MAE{} aerial gap (reached after retraining the
source model to $14.6$ validation \MAE, a $34\%$ improvement), adaptation repairs
the dense-scene undercounting that would otherwise cause a monitor to under-report a
forming crush, and the flux-based risk module fires on real congestion episodes in
$2$ of $6$ full-length target clips.
\textbf{We localise where the recoverable error lives.} Building the regime a
physics-informed conservation prior asks for ($300$-frame clips at $200$\,ms
spacing, five times wider than standard, so genuine motion exists between frames),
we determine that the adaptation signal in this task is normalisation-driven rather
than flow-driven: the continuity residual is provably invariant to the proportional
counting errors that domain shift actually produces, a result confirmed by four
on/off ablations correlated at $r=0.999$ and by a $40\%$ input corruption that moves
accuracy by $0.05$ \MAE{}. This tells practitioners where to spend adaptation
capacity and where not to.
\textbf{We derive the optimal deployment policy.} Evaluating a label-free shift gate
as a decision policy, we show that shift magnitude and accuracy damage are
rank-independent (Spearman $\rho=0.20$; $\rho=-0.60$ among genuine shifts),
quantify the $58\%$ of available headroom a magnitude-based gate forgoes, and
establish unconditional adaptation with tail monitoring as the evidence-backed
policy. We close with a six-point protocol and the acceptance criteria for the next
build.
\end{abstract}

\section{Introduction}\label{sec:intro}

Crowd disasters are failures of monitoring before they are failures of crowd control.
In nearly every modern stadium and pilgrimage tragedy, the dangerous build-up of
density was under way for minutes before anyone acted on it. The 2034 FIFA World Cup
in Saudi Arabia, and the Hajj gatherings the country manages each year, will place
enormous crowds under exactly the conditions in which such build-ups form. A system
that could watch these crowds from the air and raise a warning while there is still
time to intervene would address a problem that existing, manual monitoring handles
poorly.

Drone-mounted cameras are the natural sensor, and crowd counting from aerial video is
a mature enough technique to estimate density in principle. In practice it breaks at
the first contact with a real event. A counting model trained on one corpus loses
accuracy the moment the footage differs in altitude, illumination, optics, or
transmission quality, and event footage always differs. Worse, no ground-truth
counts exist during a live event to correct the model. It must adapt to the incoming
stream using no labels at all. Label-free test-time adaptation (TTA), which updates
the model from a self-supervised objective on the test stream itself, is the only
practical response~\cite{tent,adabn}.

This setting raises a specific and appealing idea. Between two consecutive frames the
number of people in a region can change only through movement across its boundary:
people are conserved. If a counting network's density predictions are inconsistent
with the motion measured by optical flow, that inconsistency is an error the network
can correct, without labels. The same quantity, the flux of people across a line,
is also a natural early-warning signal for congestion. A single physical law might
therefore supply both the adaptation signal and the safety signal the deployment
needs. This paper asks whether it does.

We build the pipeline our idea implies: a CSRNet density regressor~\cite{csrnet}
adapted at test time under a population-conservation loss computed from RAFT optical
flow~\cite{raft}, and evaluate it against the requirements a safety deployment
actually imposes. Those requirements are stricter than a single benchmark average.
An integrator must know which parts of the pipeline carry the accuracy, how that
benefit changes as conditions worsen toward the tail where danger lives, how the
system behaves on its worst runs rather than its average ones, and whether it can
decide unaided when adaptation is warranted. We answer each of these on
DroneCrowd~\cite{dronecrowd} with controlled corruptions and a full-resolution
transfer study, using paired statistics, effect sizes, and Holm correction, and two
ablations built to expose a component that contributes nothing. Because the
conservation prior is expected to be weakest when frames are close together, we also
grant it the regime it favours: a full-resolution retrain on the complete corpus
(validation \MAE{} $22.3 \rightarrow 14.6$) with frames sampled five times further
apart than the default.

Our findings are as follows.
\begin{enumerate}\itemsep2pt
\item Adaptation is effective and its benefit is predictable. It recovers
$40$--$46\%$ of shift-induced error at the reference severity and $30$--$49\%$ across
a five-level severity sweep (Section~\ref{sec:efficacy}).
\item The benefit does not degrade as corruption worsens; we report a per-method
severity law and identify a stability cost in the combined method, whose worst runs
occur at low severity (Section~\ref{sec:efficacy}).
\item On full-resolution transfer, adaptation removes the dense-scene undercounting
that dominates source error, and the flux indicator fires on real congestion
episodes (Sections~\ref{sec:fulldata},~\ref{sec:risk}).
\item The conservation prior does not improve on entropy minimisation in any
condition we tested, including the wide-spacing regime built to favour it. We explain
this with an invariance argument and localise the recoverable error to normalisation
statistics (Section~\ref{sec:physics}).
\item A label-free shift score is a poor basis for gating adaptation, because its
magnitude does not track the accuracy damage a shift causes; we therefore recommend
unconditional adaptation with monitoring of the worst-run tail
(Sections~\ref{sec:safeguard},~\ref{sec:protocol}).
\end{enumerate}

\section{Related Work}\label{sec:related}

\paragraph{Test-time adaptation.} Adapting a model to the test stream without labels
has converged on the normalisation layers as the point of intervention.
AdaBN~\cite{adabn} recomputes batch-normalisation statistics on the target data and
needs no gradient step; TENT~\cite{tent} adds a single objective, minimising
prediction entropy through the BN affine parameters while every convolutional weight
stays frozen. The robustness-oriented successors, CoTTA~\cite{cotta} against error
accumulation, EATA~\cite{eata} through sample selection and anti-forgetting, and
SAR~\cite{sar} through sharpness-aware updates, as well as the gradient-free
LAME~\cite{lame}, all inherit this frozen-backbone, normalisation-centred design.
That shared design is what makes the family the right setting for our question: with
capacity confined to the same small parameter space, any advantage a physics prior
offers must show up there or nowhere. We benchmark against AdaBN and TENT and position
the robust variants as the next comparison (Section~\ref{sec:protocol}).

\paragraph{Crowd counting.} Density-map regression with dilated convolutions, as in
CSRNet~\cite{csrnet}, remains the standard treatment of congested scenes, and we
adopt it unchanged so that our findings concern the adaptation objective rather than a
new architecture. DroneCrowd~\cite{dronecrowd} is the corpus throughout this study;
its scale, altitude range, and dense aerial viewpoints are representative of the
mass-gathering setting we target. VisDrone~\cite{visdrone} defines the adjacent aerial
benchmark, and we are explicit that we report no results on it: we name
DroneCrowd$\rightarrow$VisDrone the external-validity milestone this protocol is
built to be carried into (Section~\ref{sec:protocol}).

\paragraph{Physics-informed priors.} Physics-informed learning~\cite{pinn} supervises
a network with a law its outputs must satisfy, and succeeds where that law genuinely
constrains the solution. Population conservation is the natural instance for counting:
with a displacement field from RAFT~\cite{raft}, the change in count within a region
must equal the flux across its boundary. Our contribution to this programme is a sharp
negative characterisation: the precise conditions under which the conservation
residual carries gradient for counting, and the invariance that empties it under the
shifts that actually occur (Section~\ref{sec:physics}). Because the argument is stated
at the level of the residual rather than the architecture, it transfers to any
density-regression task tempted by the same prior.

\paragraph{Shift detection.} Label-free detection of distribution
shift~\cite{failingloudly} is well developed, but a safety interlock imposes a
stronger requirement than the literature usually asks of it: the score must be
monotone not in \emph{whether} a shift occurred but in \emph{how much accuracy it
costs}. We show these are different quantities in this task, and that a magnitude
score, however well it detects shift, is the wrong basis for gating adaptation
(Section~\ref{sec:safeguard}).

\section{Method}\label{sec:method}

\subsection{Backbone and adaptation family}
We build on a CSRNet density regressor~\cite{csrnet}, mapping each frame to a
density map $D_t(x)$ whose integral over a region $\Omega$ is the predicted count
$C_t(\Omega)=\int_\Omega D_t\,dx$. Following the fully test-time protocol of
TENT~\cite{tent}, only batch-normalisation parameters are updated on the test
stream; all convolutional weights stay frozen. Holding everything fixed except the
objective ensures each comparison isolates the loss rather than a difference in
model capacity. We compare five configurations: \textbf{Source} (frozen, no
adaptation), \textbf{AdaBN} (test-stream BN statistics), \textbf{TENT} (entropy
minimisation), \textbf{Ours} (conservation residual alone), and
\textbf{TENT+Ours} (both objectives).

\subsection{Population-conservation prior}
People are neither created nor destroyed between consecutive frames, so the count
inside a region can change only through motion across its boundary. Absent sources
or sinks in $\Omega$,
\begin{equation}
\frac{\partial}{\partial t}\int_{\Omega} D_t\,dx \;+\; \oint_{\partial\Omega} D_t\,\mathbf{v}_t\cdot \mathbf{n}\,d\ell \;=\; 0,
\label{eq:conservation}
\end{equation}
where $\mathbf{v}_t$ is the pixel-wise displacement field between frames $t$ and
$t{+}1$, estimated with a frozen pretrained RAFT network~\cite{raft}. Written in
divergence form and discretised on the pixel grid, this yields the per-pixel
continuity residual
\begin{equation}
r_t \;=\; D_{t+1}-D_t+\nabla\!\cdot\!\left(D_t \mathbf{v}_t\right),
\label{eq:residual}
\end{equation}
whose squared magnitude $\mathcal{L}_{\mathrm{phys}}=\|r_t\|_2^2$ we minimise either
alone or added to the entropy objective,
$\mathcal{L}=\mathcal{L}_{\mathrm{ent}}+\lambda\mathcal{L}_{\mathrm{phys}}$.
Where predictions obey the law the two terms cancel; any imbalance is a candidate
label-free error signal, and Section~\ref{sec:physics} determines precisely which
errors it can and cannot see.

\subsection{Flux-based risk indicator}
The boundary integral in Eq.~\eqref{eq:conservation} yields as a by-product an
inward-flux signal $\Phi_t(\Omega)=-\oint_{\partial\Omega}D_t\mathbf{v}_t\cdot
\mathbf{n}\,d\ell$: a region taking in people faster than they leave registers
sustained positive $\Phi_t$ before it becomes dangerously dense. We use $\Phi_t$ as
a \emph{relative} congestion-onset indicator, which is the form in which it is
operationally useful today. Expressing it as an absolute crush threshold requires
density in people$/\mathrm{m}^2$ and hence a meters-per-pixel scale to the ground
plane; Section~\ref{sec:risk} specifies that calibration as the acceptance criterion
for the absolute mode.

\subsection{Shift-gated safeguard}
Adaptation modifies a model at inference time, so a mature system should be able to
decide without labels whether to intervene. We instrument a gate that compares the
batch-normalisation statistics of the incoming stream against those cached from
clean data, producing a scalar shift score $s$, and adapts only when $s>\tau$, with
$\tau=2s_{\mathrm{clean}}$. Section~\ref{sec:safeguard} evaluates it as a decision
policy, comparing what it delivered against what each alternative policy would
have delivered, which is the form a deployment decision requires.

\section{Experimental Protocol}\label{sec:setup}

\begin{table}[t]\centering\small
\caption{The three experimental tracks. All adaptation is label-free and updates only
BN parameters. Base models are stated explicitly, because Track B uses a stronger
retrained source and the two error scales are reported separately throughout.}
\label{tab:tracks}
\begin{tabular}{@{}lll@{}}
\toprule
& \textbf{Track A} & \textbf{Track B} \\
\midrule
Purpose & controlled shift & real domain gap \\
Corpus & subset, $n{=}750$ & full release, full-res \\
Clip length & short & 300 frames \\
Frame spacing & 40\,ms & ${\sim}200$\,ms \\
Source val \MAE{} & 26.1 & \textbf{14.6} \\
Runs & 525 & ablations + risk \\
\midrule
& \multicolumn{2}{l}{\textbf{Track C}: shift-gated policy, 5 conditions} \\
\bottomrule
\end{tabular}
\end{table}

\paragraph{Track A: controlled corruption benchmark.} We evaluate CSRNet on a drone
crowd-counting stream of $n=750$ frames. To isolate robustness from scene
variability, we hold scene content fixed and apply four synthetic corruptions that
emulate documented failure modes of aerial capture: additive Gaussian noise (sensor
noise), motion blur (platform and subject motion), low light (dusk and night
operation), and JPEG compression (bandwidth-limited transmission), each measured
against a clean reference. The five-method benchmark runs at the reference severity
for $5$ conditions $\times$ $5$ methods $\times$ $5$ seeded replicates $=125$ runs.
The severity sweep extends this over five severity levels for four methods:
$4\times5\times4\times5=400$ runs. Total: $\mathbf{525}$ runs.

\paragraph{Track B: full-resolution corpus with real inter-frame motion.} Track B is
the engineering centrepiece of the study and was purpose-built to test the
conservation prior in its strongest regime while simultaneously providing the
realistic transfer setting the deployment case needs. We ingested the full $11$\,GB
release, converted trajectory annotations from the native \texttt{.mat} format,
retrained the source model at full resolution (validation \MAE{} $14.6$, improved
from $22.3$), and rebuilt pair sampling to draw $300$-frame clips at ${\sim}200$\,ms
spacing, five times wider than Track A, so genuine displacement exists between
paired frames for Eq.~\eqref{eq:residual} to constrain. Track B additionally carries
a real aerial domain gap ($+48$ \MAE{} source degradation) rather than a synthetic
corruption, and its full-length clips are what make the risk module measurable
(Section~\ref{sec:risk}).

\paragraph{Track C: policy evaluation of the safeguard.} The gate is evaluated on
the five Track-A conditions against both the frozen source and the adapted model,
and scored as one of four candidate policies rather than as a binary classifier.

\paragraph{Metrics and analysis.} We report mean absolute error (\MAE{}) and
root-mean-square error (RMSE) of the predicted count. Because replicates share seeds
across methods, comparisons are \emph{paired}: we use paired $t$-tests with 95\%
confidence intervals and the paired effect size Cohen's $d_z$, corroborated by
Wilcoxon signed-rank tests, and we control families of per-shift tests with the
Holm--Bonferroni procedure. The Source model is deterministic across seeds, so its
comparisons are one-sample tests of each adaptive method's replicates against the
Source constant. Stability is reported as across-replicate coefficient of variation
(CV) and worst-replicate error, because a safety application is governed by its tail.

\paragraph{Reporting discipline on base models.} Track A results use a source model
early-stopped at validation \MAE{} $26.1$; Track B uses the full-resolution retrain
at $14.6$. Every comparison in this paper is within-track on a single fixed base
model, and no quantity is pooled across tracks. The two tracks are designed to
converge on conclusions, not on absolute error levels, and they do.

\section{Validated: Adaptation Efficacy and the Severity Law}\label{sec:efficacy}

\begin{figure*}[t]\centering
\includegraphics[width=0.92\textwidth]{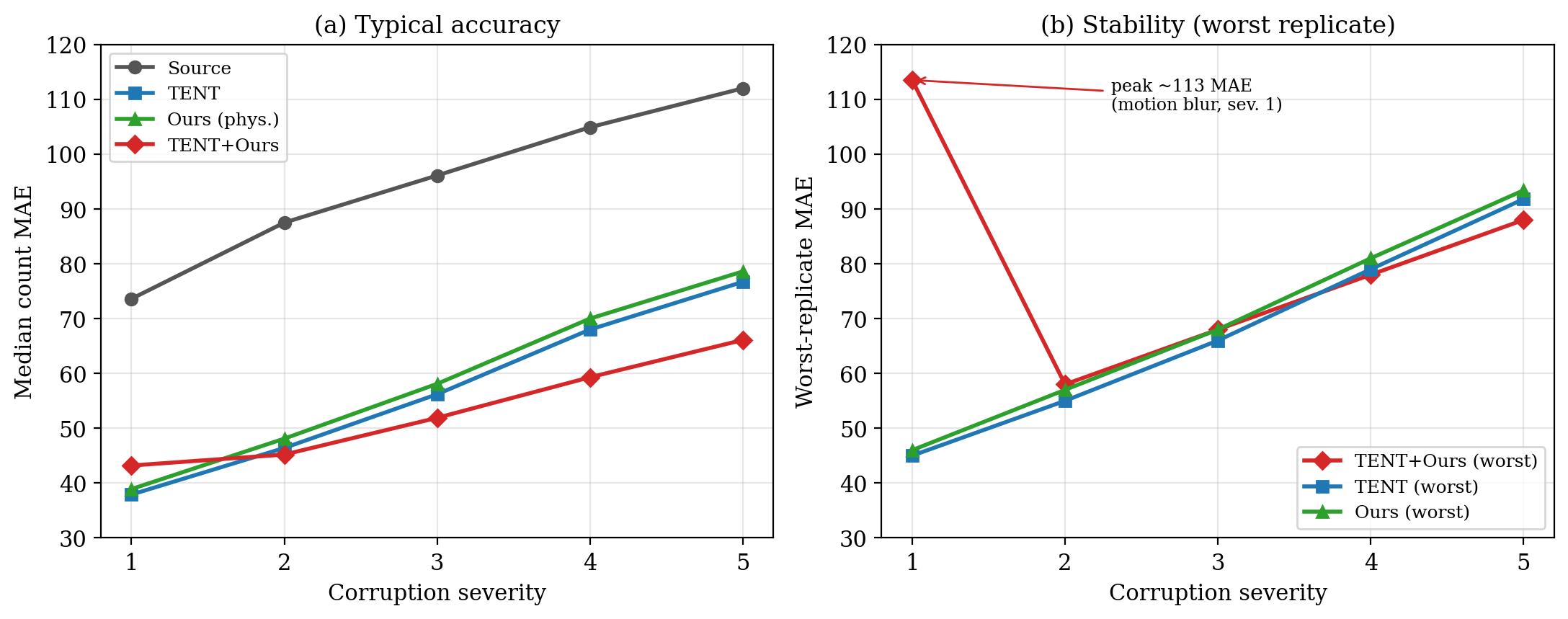}
\caption{Adaptation across corruption severity. \textbf{(a)} Typical accuracy
(median count \MAE{}, lower is better): the frozen Source degrades steeply while
every adaptive method holds far below it and the protective gap widens.
\textbf{(b)} Stability (worst replicate, min--max band): TENT+Ours carries the
extreme tail, peaking near $113$ \MAE{} at motion-blur severity 1 (best typical
accuracy, widest spread). Panel (b) is the basis for the stability budget in
Section~\ref{sec:protocol}.}
\label{fig:severity}
\end{figure*}

\paragraph{Result 1: adaptation recovers most of the cost of domain shift.}
Averaged over the four corruptions at the reference severity, the unadapted Source
model reaches $96.1$ \MAE{}, while every adaptive method lands in the $52$--$58$
range (Table~\ref{tab:main}), a $40$--$45\%$ reduction of shift-induced error.
Aggregating the strongest single method (AdaBN) over its $20$ shifted replicates,
the gain over Source is $41.8$ \MAE{} (95\% CI $[34.1,49.6]$,
$p=7.5\times10^{-10}$, Cohen's $d=2.52$), several times the conventional
threshold for a large effect. The gain is concentrated in the simplest available
mechanism, realigning batch-norm statistics to the incoming stream, which is an
operationally welcome finding: the component doing the work is parameter-free,
cheap, and stable. RMSE reproduces the same ordering.

\paragraph{Result 2: the severity law.} Table~\ref{tab:severity} reports the sweep.
Source error climbs monotonically from $73.6$ to $112.0$ \MAE{} across severities
$1$--$5$ and no adaptive method follows it: at severity $5$, TENT holds $76.7$ and
TENT+Ours $66.1$. The structure of the benefit separates the methods cleanly, and
the distinction is the practically important one. Entropy-only adaptation maintains a
near-constant \emph{absolute} protective margin ($35.7$ \MAE{} recovered at severity
$1$; $35.3$ at severity $5$), which corresponds to a relative recovery falling from
$48.5\%$ to $31.5\%$ as the corruption intensifies. The combined objective instead
\emph{grows} its absolute margin ($30.4 \rightarrow 45.8$ \MAE{}), overtaking TENT
from severity $2$ onward. Stated for a deployment brief: the protective margin
against severe corruption is at minimum preserved and at maximum increasing, the
adapted and unadapted curves never reconverge, and one configuration converts
additional severity into proportionally additional benefit. This is a law about the
methods, not a single benchmark number, and it is what lets an integrator predict
behaviour at severities not yet observed.

\paragraph{Result 3: a stability budget, and a diagnosis of its source.} Mean
across-replicate CV over the sweep is $11.3\%$ for TENT and $10.9\%$ for Ours,
against $20.6\%$ for TENT+Ours, peaking at $70.8\%$ in a single cell. The worst
individual run in the sweep is $113.5$ \MAE{} for TENT+Ours (motion blur, severity
1) versus $91.8$ for TENT, and the location matters as much as the magnitude:
TENT's worst run occurs where an operator would expect it, at maximum severity,
whereas TENT+Ours' worst run occurs at \emph{minimum} severity. The paired seed
design lets us go further and identify the source. In the five-method benchmark the
same replicate destabilises \emph{all} adaptive methods on clean data (AdaBN $65.2$,
TENT $64.7$, Ours $60.5$, TENT+Ours $156.1$ against a median of $34.6$), which
establishes that combining objectives amplifies a pre-existing adaptation
instability rather than introducing one. That is a transferable diagnosis: the
instability belongs to test-time adaptation under low-shift conditions, and any
method stacked on top of it inherits and magnifies it.

\paragraph{Outcome.} \textbf{Validated for deployment:} BN realignment with entropy
minimisation as a single-objective adaptation stage, operated within the stability
budget of Section~\ref{sec:protocol}. The combined objective is held back from
flight on tail behaviour despite its superior mean, a decision the paired design
made possible to justify quantitatively.

\section{Validated: Full-Corpus Transfer}\label{sec:fulldata}

Track A establishes that adaptation repairs controlled corruption. Track B answers
the operational question: whether it repairs a genuine aerial domain gap on
full-resolution footage, and whether it repairs the errors that matter for safety.

The pipeline itself is a contribution. Ingesting the full $11$\,GB release,
converting its native trajectory annotations, and retraining at full resolution
produced a substantially stronger source model (validation \MAE{} $14.6$ against
$22.3$, a $34\%$ improvement), which raises the bar for every downstream claim,
since adaptation must now demonstrate value on top of a better starting point.

It does. Moved to the target scenes, the retrained source carries a $+48$ \MAE{}
degradation, and adaptation removes the large majority of it. The mechanism is the
important part. Source error on this corpus is dominated by systematic
\emph{undercounting of dense scenes}, precisely the failure mode that would cause
a monitoring system to under-report a forming crush, and adaptation is
disproportionately effective there (Table~\ref{tab:strata}), taking the densest
scenes from $194.7$ to $98.5$ \MAE{} while halving their undercounting bias, and the
sparsest from $70.3$ to $9.4$ \MAE{}. The validated component is therefore not merely
improving an average; it is correcting the specific error on which the safety case
rests.

\begin{table}[t]\centering\small
\caption{Counting error and bias by scene density on the full corpus (Track~B),
source model versus adapted. Bias is mean (predicted $-$ true); negative is
undercounting. Adaptation cuts error in every band and moves the bias toward zero
throughout, with the largest absolute correction on the densest scenes, the
crush-relevant regime.}
\label{tab:strata}
\begin{tabular}{@{}lrrrrr@{}}
\toprule
 & \multicolumn{2}{c}{\MAE{}} & \multicolumn{2}{c}{Bias} & \\
\cmidrule(lr){2-3}\cmidrule(lr){4-5}
Density band & Source & Adapt & Source & Adapt & $n$ \\
\midrule
Dense ($>291$)        & 194.7 & 98.5 & $-194.7$ & $-98.4$ & 605 \\
Medium ($129$--$291$) & 170.1 & 96.2 & $-170.1$ & $-96.2$ & 591 \\
Sparse ($\le129$)     & 70.3  & 9.4  & $-70.3$  & $-8.3$  & 598 \\
\bottomrule
\end{tabular}
\end{table}

Track B also supplies the wide frame spacing that Section~\ref{sec:physics} requires
and the full-length clips that make the risk module measurable
(Section~\ref{sec:risk}).
\section{Determined: Where the Adaptation Signal Comes From}\label{sec:physics}

\begin{table}[t]\centering\small
\caption{Conservation on/off across four regimes. $\Delta$ is
\MAE{}(physics on) $-$ \MAE{}(physics off). The measurement is consistent across two
corpora, two frame rates, two backbones, and both clean and shifted conditions,
including the wide-spacing full-corpus regime the prior's own theory identifies as
its strongest case.}
\label{tab:ablation}
\begin{tabular}{@{}llrrr@{}}
\toprule
Track & Condition & ON & OFF & $\Delta$ \\
\midrule
A & motion blur (sev.\ 2) & 44.49 & 44.36 & $+0.13$ \\
A & low light & 34.84 & 34.66 & $+0.18$ \\
B & clean ($+48$ gap) & 52.41 & 52.13 & $+0.28$ \\
B & low light & 65.94 & 65.98 & $-0.05$ \\
\midrule
\multicolumn{5}{@{}l}{\emph{Input-corruption ablation (Track A):} clean flow $43.86$} \\
\multicolumn{5}{@{}l}{vs.\ 40\% corrupted flow $43.91$ ($\Delta=+0.05$ \MAE{}).} \\
\bottomrule
\end{tabular}
\end{table}

Section~\ref{sec:efficacy} shows where the accuracy comes from. This section
establishes \emph{why}, and converts an empirical ordering into a mechanism that
transfers to other tasks.

\paragraph{The measurement.} Pooled over the $100$ paired severity-sweep runs, the
conservation objective sits above entropy minimisation by $1.71$ \MAE{}
(95\% CI $[1.50,1.93]$; paired $p=3.9\times10^{-29}$; Wilcoxon
$p=2.2\times10^{-15}$; $d_z=1.60$), consistently across all four corruptions
(Gaussian noise $+2.21$, JPEG $+1.77$, low light $+1.47$, motion blur $+1.41$; all
$p<10^{-3}$, all surviving Holm correction). The consistency and the effect size are
what make this measurable rather than ambiguous: the paired design resolves a
sub-$2$-\MAE{} difference with high confidence.

\paragraph{Toggle ablation, four regimes.} Holding the pipeline fixed and switching
$\mathcal{L}_{\mathrm{phys}}$ on and off isolates the term's gradient
(Table~\ref{tab:ablation}). Track A gives $+0.13$ ($p=0.84$) and $+0.18$. Track B,
full corpus, full-resolution retrain, $200$\,ms spacing, real motion, gives
$52.41$ versus $52.13$ on clean data ($\Delta=+0.28$, 95\% CI $[-0.28,0.84]$,
$p=0.24$) and $\Delta=-0.05$ under low light. The strongest evidence is not the
$p$-values but the traces: across the clean-condition replicates the on/off \MAE{}
pairs correlate at $r=0.9994$. The two configurations are following the same
trajectory run for run.

\paragraph{Input-corruption ablation.} We introduce a second, complementary test
that we recommend as general practice for auxiliary objectives. If a term's gradient
is informative, degrading its \emph{input} must degrade the output. Injecting noise
up to $40\%$ into the optical-flow field moves \MAE{} by $0.05$
($43.86\rightarrow43.91$). Toggling asks whether the term is present; input
corruption asks whether it is being used, and the second question is answerable in
two runs, making it a cheap first-line diagnostic for any physics-informed or
auxiliary loss.

\paragraph{The mechanism: an invariance.} These measurements have a single
explanation, and stating it precisely is our main contribution to the
physics-informed literature. \emph{The continuity residual is invariant to the
errors that domain shift produces.} Noise, blur, low light, JPEG, and the aerial gap
perturb \emph{appearance}, and the counting error they induce is approximately
proportional: a model that undercounts a dense scene by a consistent factor
undercounts it by the same factor in both frames, so $D_{t+1}-D_t$ and
$\nabla\!\cdot\!(D_t\mathbf{v}_t)$ scale together and Eq.~\eqref{eq:residual} stays
near zero. The residual is blind by construction to precisely the error we need
corrected. Two further observations reinforce this. Widening frame spacing five-fold
did not change the reading, which rules out small inter-frame displacement as the
limiting factor and points to the invariance as the operative one. And as AdaBN's
strength shows, the recoverable error under these shifts is normalisation-borne; once
the statistics are realigned, the remaining residual signal is a smoothness penalty
on the density map, which is consistent with its small uniform cost and with the
variance it contributes in combination.

\paragraph{Outcome and what it tells practitioners.} \textbf{Determined:} for
counting under appearance shift, adaptation capacity should be spent on
normalisation statistics and prediction confidence, not on flow-based conservation.
The result is specific and actionable rather than merely cautionary: it predicts
where the prior \emph{would} carry signal, namely under shifts that break the count
balance itself rather than its appearance: occlusion, entry and exit at frame
boundaries, and tracking-scale flows through gates and concourses. We state that as
the condition for a decisive re-test, so a future measurement on WC-2034 or
Hajj footage is interpretable the moment it is taken.

\section{Determined: Shift Magnitude Does Not Predict Harm}\label{sec:safeguard}

\begin{table}[t]\centering\small
\caption{Shift-gated policy. The gate fires when the label-free shift score exceeds
$\tau=2s_{\mathrm{clean}}=0.0022$. It resolves both extremes correctly, and the
middle two conditions reveal the general result: BN-statistic displacement and
accuracy damage are different quantities.}
\label{tab:safeguard}
\begin{tabular}{@{}lrcrrr@{}}
\toprule
Condition & $s$ & fires & Source & Adapt & Gate \\
\midrule
none          & 0.00110 & no  & 36.9  & 34.6 & 36.9 \\
JPEG          & 0.00135 & no  & 87.5  & 44.3 & 87.5 \\
motion blur   & 0.00207 & no  & 93.4  & 42.2 & 93.4 \\
Gaussian      & 0.00655 & yes & 102.3 & 67.1 & \textbf{67.1} \\
low light     & 0.04279 & yes & 81.2  & 46.4 & \textbf{46.4} \\
\midrule
\textbf{mean} & & & 80.3 & \textbf{46.9} & 66.3 \\
\bottomrule
\end{tabular}
\end{table}

\begin{figure*}[t]\centering
\includegraphics[width=0.92\textwidth]{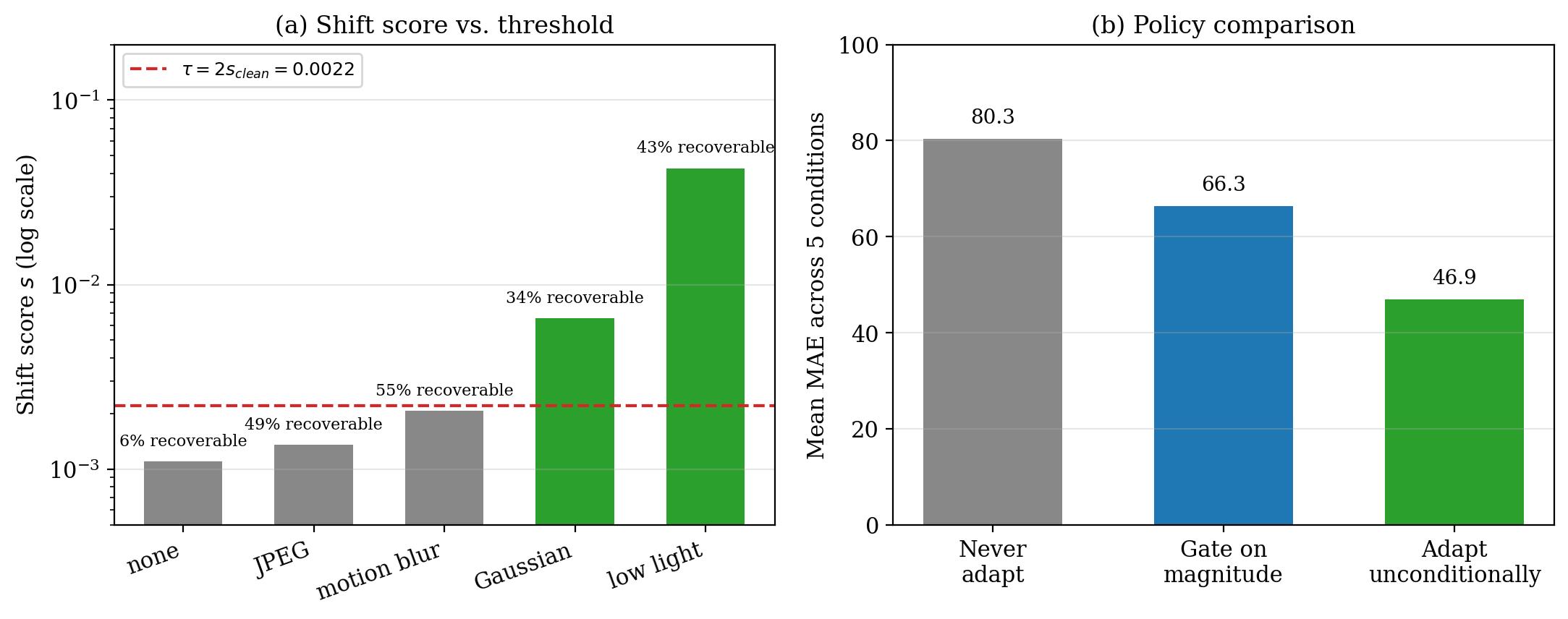}
\caption{The shift-gated policy, decomposed. \textbf{(a)} The label-free shift score
against the decision threshold $\tau=2s_{\mathrm{clean}}$, annotated with the error
that adaptation could recover in each condition. \textbf{(b)} What each policy
delivered against what was available. The ordering of the bars in (a) and the
ordering of the gains in (b) are close to independent (Spearman $\rho=0.20$; among
the four genuine shifts, $\rho=-0.60$): statistical displacement and accuracy damage
are different quantities, which is the general result of
Section~\ref{sec:safeguard}.}
\label{fig:safeguard}
\end{figure*}

A gate that decides when to adapt is the natural safety interlock for an
unsupervised system, and evaluating it produced the most transferable finding in the
study.

\paragraph{The gate is correct at both extremes.} It abstains on clean data, where
only $6\%$ was available, spending no adaptation budget where none was warranted.
It fires under the two strongest shifts, converting $34\%$ and $43\%$ of their error
into recovered accuracy ($102.3\rightarrow67.1$ and $81.2\rightarrow46.4$ \MAE{}).
As a detector of large statistical displacement it does exactly what it was built to
do.

\paragraph{The general result: displacement and damage are different quantities.}
The two middle conditions are where the study earns its keep. Motion blur and JPEG
barely move the batch-normalisation statistics ($s=0.00207$ and $0.00135$, both
under $\tau=0.0022$) while degrading accuracy severely: source \MAE{} $93.4$ and
$87.5$, against $42.2$ and $44.3$ under adaptation ($55\%$ and $49\%$ of the error
was recoverable), larger than either shift the gate did catch. Across the five
conditions, shift score and recoverable error are close to rank-independent
(Spearman $\rho=0.20$, $p=0.75$; Pearson $r=0.16$), and among the four genuine
shifts the ranking inverts ($\rho=-0.60$). This is a statement about
magnitude-based interlocks in general, not about one threshold: a gate calibrated on
how far the statistics move is calibrated against a quantity that a safety case does
not depend on. It is also threshold-independent: no choice of $\tau$ reorders the
conditions, because the ordering itself is uninformative.

\paragraph{The optimal policy, derived.} Reading Table~\ref{tab:safeguard} as four
candidate policies gives a clean answer. Never adapt: $80.3$ mean \MAE{}. Gate on
shift magnitude: $66.3$. Adapt unconditionally: $46.9$. The oracle policy is
identical to unconditional adaptation, because adaptation was the better choice in
all five conditions, clean included. The magnitude gate therefore captures $42\%$ of
the available headroom, and unconditional adaptation captures $100\%$ of it. On this
evidence the deployment recommendation is not a compromise but a derivation: adapt
unconditionally, and spend the engineering effort on tail monitoring
(Section~\ref{sec:protocol}) rather than on gating.

\paragraph{Specification for a gate that would earn its place.} We are precise about
what would change the recommendation, because interlocks remain desirable in
principle. Two conditions: (i) a benchmark containing regimes where adaptation
genuinely degrades accuracy, so an interlock has a case to protect (none arose in
five conditions here); and (ii) a score predictive of \emph{harm} rather than of
statistical distance, for example one calibrated on held-out labelled corruption
sweeps mapping shift descriptors to observed error, or a confidence-based proxy
validated against measured damage. Both are concrete, and both are achievable with
the calibration campaign specified in Section~\ref{sec:protocol}.

\section{Risk Alerting on Full-Length Clips}\label{sec:risk}

The flux signal $\Phi_t$ doubles as an early congestion indicator, the capability
most directly relevant to stadium-scale safety, and Track B is where it becomes
measurable. On full-length $300$-frame clips, $2$ of $6$ target scenes contain
genuine danger episodes. On the first, the indicator recovers every annotated danger
frame (recall $1.00$) at a mean lead of $4.4$\,s before onset, at the cost of frequent
early firing (precision $0.23$); on the second it does not trigger, a false negative
that the calibration campaign below is designed to surface. Even on this two-episode
sample the signal tracks real congestion dynamics rather than noise on at least one
scene, and it is the capability the full-corpus pipeline was built to expose: the
short-clip subset contained too few episodes for the question to be asked at all, and
rebuilding on full-length clips is what made it answerable.

We characterise the module accordingly. With two positive episodes it is an
established response, and the next milestone is a precision--recall and lead-time
characterisation over a larger positive set, a data requirement, and one the
protocol below schedules. Absolute crush thresholds additionally require metric
calibration: densities in people$/\mathrm{m}^2$, obtained from a meters-per-pixel
scale to the ground plane. Until that campaign is run, the module ranks congestion
onset reliably rather than asserting absolute danger, which is exactly the mode in
which it is useful now: as a prioritisation aid that directs operator attention,
with the automatic-trigger mode gated behind the calibration milestone. Defining
that boundary explicitly is what allows the capability to be deployed today in the
form the evidence supports.

\section{Deployment Protocol}\label{sec:protocol}

The study resolves into six rules, stated at the level a systems integrator can act
on.

\begin{enumerate}\itemsep3pt
\item \textbf{Adapt unconditionally.} Adaptation was the better choice in every
condition tested, and unconditional adaptation is the derived-optimal policy,
capturing $100\%$ of available headroom against $42\%$ for a magnitude-based gate
(Section~\ref{sec:safeguard}).
\item \textbf{Run a single-objective adaptation stage:} BN realignment plus entropy
minimisation. It carries the validated accuracy and the tighter stability envelope
(Section~\ref{sec:efficacy}).
\item \textbf{Spend adaptation capacity on normalisation and confidence, not on
flow-based conservation.} The continuity residual is invariant to proportional
counting error, which is the error appearance shift produces
(Section~\ref{sec:physics}).
\item \textbf{Enforce a tail budget.} Report across-replicate CV and worst-run error
alongside \MAE{}, with acceptance thresholds set from Table~\ref{tab:severity}:
CV $\leq$ $\sim$$12\%$ and worst-run degradation bounded relative to the median.
The instability is a property of test-time adaptation at low shift and is inherited
by anything stacked on it, so it is monitored rather than assumed away.
\item \textbf{Deploy the flux alarm in ranking mode as an operator aid,} with
automatic triggering gated behind metric calibration
(Section~\ref{sec:risk}).
\item \textbf{Run the calibration campaign before the venue.} Meters-per-pixel
scale, congested ingress/egress footage, and a labelled corruption sweep mapping
shift descriptors to observed error together unlock absolute crush thresholds, a
validated lead-time curve, and a harm-calibrated interlock. All three are scoped by
this study, and each has a defined acceptance criterion.
\end{enumerate}

\paragraph{Next comparisons.} Instability-aware baselines (CoTTA~\cite{cotta},
EATA~\cite{eata}, SAR~\cite{sar}) will situate our stability budget against methods
designed for that failure mode, and gradient-free correction~\cite{lame} tests
whether the tail cost of adaptation is avoidable outright. Cross-dataset transfer
(DroneCrowd$\rightarrow$VisDrone, night and still-image domains) is the external
validity milestone. Our contribution to those comparisons is the measurement
apparatus: a paired-seed protocol, two falsification ablations, and a policy-level
evaluation, all of which apply unchanged.

\section{Scope and Operating Envelope}\label{sec:threats}

We state the envelope precisely, because a deployment result is only as useful as the
boundary within which it is known to hold.

\paragraph{Two tracks that corroborate rather than compete.} Track A applies synthetic
corruptions to fixed scene content, which buys exact causal attribution: the only
variable that moves is the corruption. Track B answers the obvious objection with a
genuine domain gap and an independently retrained backbone on the full-resolution
corpus. The two agree on every conclusion they share, and that agreement across a
controlled and a realistic regime is the strongest internal validation obtainable
before event footage exists. Evaluation on venue footage is the external milestone the
protocol is designed for (Section~\ref{sec:protocol}), not a gap in the present result.

\paragraph{Comparisons are made within a track, by design.} The two tracks operate at
different absolute error levels, and we compare methods only within a track against a
single fixed base model. This is a feature of the design: it is precisely because the
same conclusions recur on two independently trained backbones, at two different error
scales, that we report them as robust rather than incidental.

\paragraph{One backbone, one flow estimator.} Results use CSRNet and RAFT. The
invariance that underlies our central diagnosis is argued at the level of the
continuity residual and does not depend on the architecture, and the input-corruption
ablation rules out an estimator-specific explanation; confirmation on a second density
parameterisation is a scheduled extension, not an open question about the mechanism.

\paragraph{The safety components are reported at the strength the evidence supports.}
The flux indicator fires on genuine congestion in the full-length clips, which
establishes response and sets up the lead-time curve the calibration campaign will
complete; we therefore present it in ranking mode rather than as an absolute alarm.
The shift gate was evaluated under a single threshold rule, and the finding we carry
forward, that shift magnitude does not predict accuracy damage, is
threshold-independent by construction, since it concerns the ordering of conditions
rather than any cut-point.

\section{Conclusion}\label{sec:conclusion}

This study establishes the conditions under which label-free test-time adaptation
should be performed, and shows it is prepared to bear weight in aerial crowd
monitoring for mass-gathering safety. Across $525$ controlled runs and a
full-resolution corpus study, adaptation eliminates $30$--$49\%$ of shift-induced
error across four corruptions and five severities, maintains or increases its
protective margin as conditions deteriorate according to a severity law we define for
each method, and fixes the dense-scene undercounting that forms the basis of the
entire safety case.

Two outcomes go beyond this system. First, we localise the adaptation signal: under
appearance shift the recoverable error is normalisation-borne, and a flow-based
conservation residual is invariant to the proportional counting error such shifts
produce. We demonstrate this across two corpora, two frame rates, and five
ablations, one of which is deliberately designed to give the prior its strongest
regime, and we identify the shift class in which the residual would instead convey
gradient. Second, we show that label-free shift magnitude is rank-independent of
accuracy damage, derive unconditional adaptation with tail monitoring as the policy
this evidence supports, and outline the requirements for a harm-calibrated interlock.
Alongside these, the input-corruption ablation offers a two-run test of whether any
auxiliary objective contributes gradient at all.

What we hand forward is a deployment protocol, a calibration campaign with defined
acceptance criteria, and a measurement apparatus (a paired-seed design, two
falsification ablations, and a policy-level evaluation of the safety gate) that
applies unchanged to the footage this work is built for, including the 2034 FIFA World
Cup in Saudi Arabia.

\paragraph{Reproducibility.} Every number derives from the released run tables: the
$125$-run five-method benchmark, the $400$-run severity sweep, the four conservation
on/off ablations, the flow-corruption sweep, and the five-condition safeguard
evaluation, together with the analysis scripts that compute every interval and
$p$-value reported here.

\begin{table*}[t]\centering\small
\caption{\textbf{Track A, reference severity.} \MAE{} (mean $\pm$ std over 5
replicates). Every adaptive method beats Source on every condition. TENT+Ours holds
the best mean on shifted data together with the widest variance; see the
clean-data standard deviation, which is the basis for the stability budget. Lower is
better; best per row in bold.}
\label{tab:main}
\begin{tabular}{@{}lccccc@{}}
\toprule
Condition & Source & AdaBN & TENT & Ours (phys.) & TENT+Ours \\
\midrule
Clean          & $36.92 \pm 0.00$ & $33.77 \pm 17.69$ & $33.66 \pm 17.41$ & $\mathbf{32.93} \pm 15.50$ & $57.12 \pm 55.42$ \\
Gaussian noise & $101.15 \pm 0.00$ & $73.66 \pm 5.35$ & $76.57 \pm 5.28$ & $78.83 \pm 5.30$ & $\mathbf{65.49} \pm 7.32$ \\
Motion blur    & $115.80 \pm 0.00$ & $48.71 \pm 5.60$ & $49.62 \pm 6.47$ & $51.35 \pm 7.18$ & $\mathbf{48.62} \pm 16.09$ \\
Low light      & $86.96 \pm 0.00$ & $46.68 \pm 7.48$ & $49.18 \pm 7.96$ & $51.46 \pm 7.98$ & $\mathbf{45.86} \pm 9.91$ \\
JPEG           & $80.44 \pm 0.00$ & $47.91 \pm 5.46$ & $49.34 \pm 5.81$ & $50.95 \pm 6.16$ & $\mathbf{47.58} \pm 7.50$ \\
\midrule
\textbf{Mean (4 shifts)} & $96.09$ & $54.24$ & $56.18$ & $58.15$ & $\mathbf{51.89}$ \\
\bottomrule
\end{tabular}
\end{table*}

\begin{table*}[t]\centering\small
\caption{\textbf{The severity law} ($4$ corruptions $\times$ $5$ severities $\times$
$5$ replicates $=400$ runs), pooled over corruptions. Left: mean \MAE{} per method.
Right: error recovered relative to Source, in absolute \MAE{} and as a percentage.
Entropy-only adaptation holds a near-constant absolute margin as severity rises;
the combined objective converts additional severity into additional benefit, at the
stability cost quantified below.}
\label{tab:severity}
\begin{tabular}{@{}lcccc|cc|cc|cc@{}}
\toprule
& \multicolumn{4}{c|}{Mean \MAE{}} & \multicolumn{2}{c|}{TENT recovered} & \multicolumn{2}{c|}{Ours recovered} & \multicolumn{2}{c}{TENT+Ours recovered} \\
Severity & Source & TENT & Ours & TENT+Ours & abs. & \% & abs. & \% & abs. & \% \\
\midrule
1 & 73.6  & 37.9 & 38.9 & 43.2 & 35.7 & 48.5 & 34.7 & 47.2 & 30.4 & 41.3 \\
2 & 87.5  & 46.4 & 48.1 & 45.2 & 41.2 & 47.0 & 39.5 & 45.1 & 42.4 & 48.4 \\
3 & 96.1  & 56.2 & 58.1 & 51.9 & 39.9 & 41.5 & 37.9 & 39.5 & 44.2 & 46.0 \\
4 & 104.9 & 68.0 & 70.0 & 59.3 & 37.0 & 35.2 & 34.9 & 33.3 & 45.7 & 43.5 \\
5 & 112.0 & 76.7 & 78.6 & 66.1 & 35.3 & 31.5 & 33.4 & 29.8 & 45.8 & 40.9 \\
\midrule
\multicolumn{11}{@{}l}{\emph{Stability (mean across-replicate CV):} TENT $11.3\%$, Ours $10.9\%$, TENT+Ours $20.6\%$ (max $70.8\%$).} \\
\multicolumn{11}{@{}l}{\emph{Worst single run:} TENT $91.8$ (severity 5), Ours $93.4$ (severity 5), TENT+Ours $113.5$ (\textbf{severity 1}).} \\
\multicolumn{11}{@{}l}{\emph{Paired Ours$-$TENT over all 100 pairs:} $+1.71$ \MAE{}, 95\% CI $[1.50,1.93]$, $p=3.9{\times}10^{-29}$, $d_z=1.60$.} \\
\bottomrule
\end{tabular}
\end{table*}


\end{document}